\documentclass[conference]{IEEEtran}
\usepackage{times}

\usepackage[numbers]{natbib}
\usepackage{multicol}
\usepackage[bookmarks=true]{hyperref}

\usepackage[utf8]{inputenc}
\usepackage{amsmath,amssymb,stmaryrd,mathtools}
\usepackage{amsfonts}

\usepackage{acronym}
\usepackage{verbatim}
\usepackage{booktabs}
\usepackage{siunitx}
\usepackage{graphics}
\usepackage{graphicx,caption}
\usepackage{flushend}
\usepackage{suffix}
\usepackage{xstring}
\usepackage{xparse}
\usepackage{expl3}
\usepackage{mathrsfs}
\usepackage{tabularx}
\usepackage{makecell}
\usepackage{array}
\usepackage{cleveref}
\usepackage{multirow}
\usepackage{diagbox}
\usepackage{rotating}
\usepackage{algorithm}
\usepackage[noend]{algpseudocode}

\usepackage{subcaption}
\usepackage{wrapfig}

\usepackage{tikz}
\usetikzlibrary{arrows,backgrounds,calc}
\usepackage{relsize}
\usepackage{float}
\usepackage{kantlipsum} 
\usepackage{lipsum}
\usepackage{stfloats}
\usepackage{siunitx}

\usepackage{todonotes}
\usepackage{soul}
\definecolor{smoothgreen}{rgb}{0.7,1,0.7}
\sethlcolor{smoothgreen}

\RequirePackage{luatex85}
\usepackage{pgfplots}
\pgfplotsset{compat=newest}
\pgfplotsset{every axis legend/.append style={%
		cells={anchor=west}}
}
\usepgfplotslibrary{polar}
\usetikzlibrary{arrows}
\tikzset{>=stealth'}

\definecolor{C1}{rgb}{0.0, 0.447, 0.741}
\definecolor{C1_light}{rgb}{0.0, 0.6032388663967612, 1.0}
\definecolor{C2}{rgb}{0.85, 0.325, 0.098}
\definecolor{C3}{rgb}{0.929, 0.694, 0.125}
\definecolor{C4}{rgb}{0.494, 0.184, 0.556}
\definecolor{C5}{rgb}{0.466, 0.674, 0.188}
\definecolor{C6}{rgb}{0.301, 0.745, 0.933}
\definecolor{C7}{rgb}{0.635, 0.078, 0.184}

\usepgfplotslibrary{groupplots}

\usetikzlibrary{shapes.geometric, arrows}

\tikzstyle{startstop} = [rectangle, rounded corners, minimum width=2cm, minimum height=1cm,text centered, draw=black, fill=none]
\tikzstyle{arrow} = [thick,->,>=stealth]

\begin{document}

\title{Structured Graph Network for Constrained Robot Crowd Navigation with Low Fidelity Simulation}

\author{
\textbf{Shuijing Liu}, 
\textbf{Kaiwen Hong}, 
\textbf{Neeloy Chakraborty}, 
\textbf{Katherine Driggs-Campbell} \\
{University of Illinois, Urbana-Champaign} \\
\texttt{\{sliu105, kaiwen2, neeloyc2, krdc\}@illinois.edu}}



%

\maketitle

\begin{abstract}
We investigate the feasibility of deploying reinforcement learning (RL) policies for constrained crowd navigation using a low-fidelity simulator. We introduce a representation of the dynamic environment, separating human and obstacle representations. Humans are represented through detected states, while obstacles are represented as computed point clouds based on maps and robot localization. This representation enables RL policies trained in a low-fidelity simulator to deploy in real world with a reduced sim2real gap. Additionally, we propose a spatio-temporal graph to model the interactions between agents and obstacles. Based on the graph, we use attention mechanisms to capture the robot-human, human-human, and human-obstacle interactions. Our method significantly improves navigation performance in both simulated and real-world environments. Video demonstrations can be found at \url{https://sites.google.com/view/constrained-crowdnav/home}.
\end{abstract}

\IEEEpeerreviewmaketitle

\section{Introduction}
\label{sec:intro}
To co-exist and collaborate with people seamlessly, robots must navigate through dynamic environments with both moving agents and static obstacles. Such environments are usually constrained: in indoor environments, walls and furniture are common, while in outdoor environments, robots must stay on their lanes or sidewalks.

Robot crowd navigation has received much attention since last century~\cite{fox1997dynamic,hoy2015algorithms, savkin2014seeking,xie2023drlvo,mavrogiannis2023winding}. 
Model-based approaches have explored various mathematical models, such as velocity obstacles~\cite{van2011reciprocal, van2008reciprocal}, dynamic windows~\cite{fox1997dynamic}, and forces~\cite{helbing1995social}, to optimize robot actions. 
Although these models consider both humans and obstacles, they are prone to failures such as the freezing problem due to unrealistic assumptions on human behaviors~\cite{trautman2010unfreezing, liu2020decentralized}.
In addition, the hyperparameters are sensitive to different environments and thus need to be hand-tuned to ensure good performance~\cite{long2018towards}.

Another line of work trains more expressive crowd navigation policies using imitation learning and reinforcement learning (RL)~\cite{chen2017decentralized,chen2019crowd,liu2020decentralized,liu2023intention,mun2023occlusion}.
However, these works focus on navigation in open spaces and ignore obstacles
and constraints.
To address this issue, other learning-based approaches use raw sensor images or point clouds to represent environments~\cite{perez2021robot,dugas2022navdreams,pokle2019deep,xie2023drlvo}. 
These end-to-end (e2e) pipelines have made promising progress with very few assumptions about the environment.  
However, to deploy the learned policies to the real world, these e2e methods need either a large amount of demonstration data from the real world or a high-fidelity simulator, both of which are expensive to obtain and prone to domain shifts between training and testing scenarios~\cite{mavrogiannis2023core,raj2024rethinking}. 

\begin{figure}[t]
\centering
\includegraphics[scale=0.45]{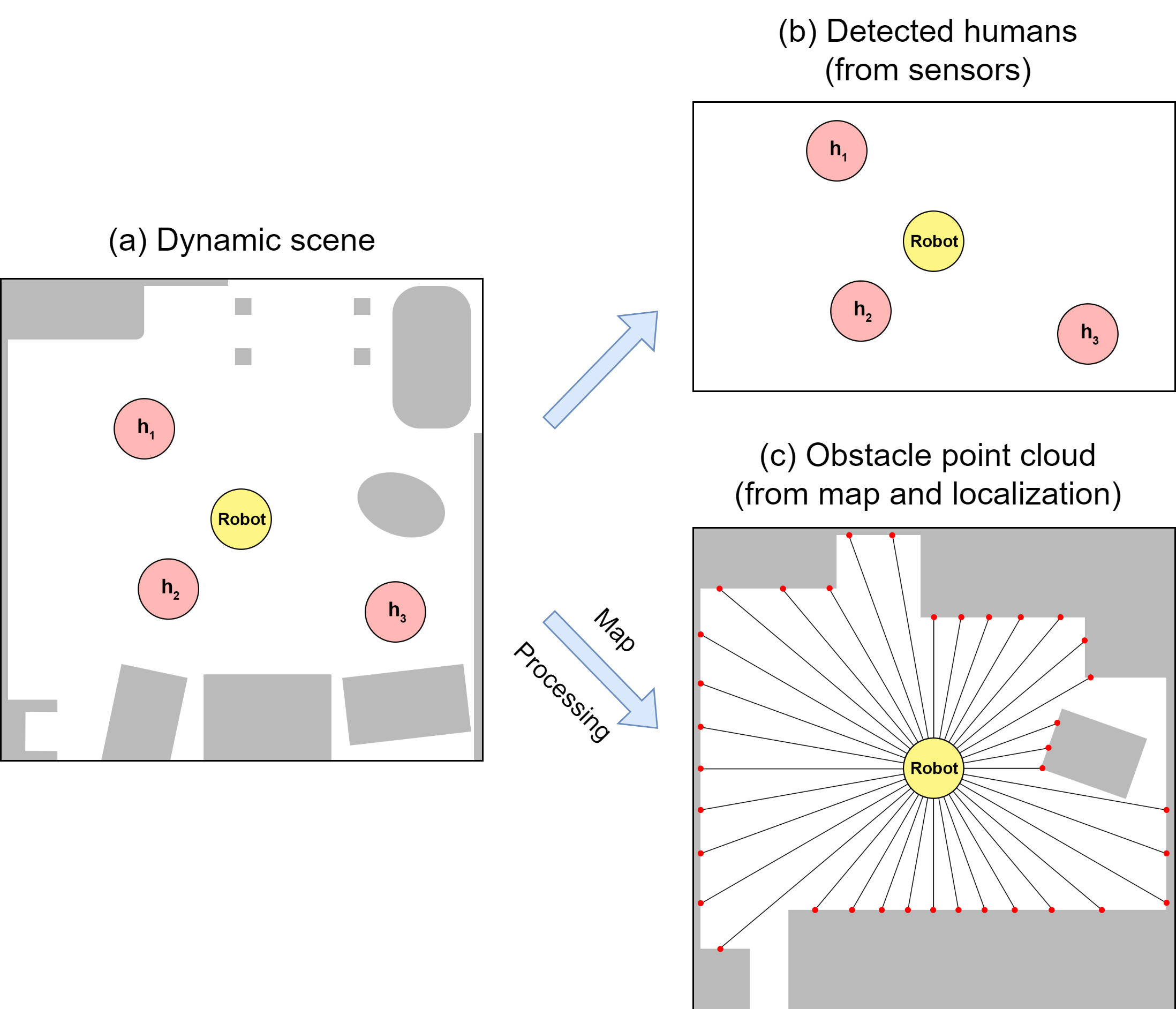}
\caption{\textbf{A split representation of constrained navigation scenario.} In a dynamic scene, human information is obtained from detections by sensors. For obstacle information, we remove all humans and compute a point cloud from a known map and the robot's location. In this way, we can learn a robot policy with smaller sim2real gaps with a cheap low-fidelity simulator.}
\label{fig:scene_rep}
\vspace{-20pt}
\end{figure}

In this paper, we ask the following question: Is it possible to deploy an RL policy for constrained crowd navigation with a cheap and low-fidelity simulator? 
The first step is to come up with an environment representation that is robust to sim2real gaps from perception. 
As shown in Fig.~\ref{fig:scene_rep}, we propose to split the human and obstacle representation and leverage on processed inputs instead of raw sensor inputs. 
We represent humans with detected states and obstacles as computed point clouds from map and robot localization.
With processed states of humans and objects, the split representation is less affected by inaccurate simulations of human gaits, visual appearance, and 3D shapes. As a result, we can train the robot policy in a low-fidelity simulator, such as the one in Fig.~\ref{fig:simulator}, with a much smaller sim2real gap compared with previous e2e methods.

With the proposed scene representation,  
we learn a robot policy that reasons about interactions among different entities. 
In constrained environments, agents have limited traversable spaces and thus interact with each other as well as obstacles frequently. 
To capture these subtle interactions, we propose spatio-temporal (st) graph and derive a novel policy network from the st-graph. We use three separate attention networks to address the different effects of robot-human, human-human, and human-obstacle interactions. After training, the attention networks enable the robot to pay more attention to important interactions, which ensures good performance when the number of humans increases and the landscape becomes complex. 


\section{Preliminaries}
\label{sec:prelim}
\subsection{MDP formulation}
We model the constrained crowd navigation scenario as a Markov Decision Process (MDP), defined by the tuple $ \langle \mathcal{S}, \mathcal{A}, \mathcal{P}, R, \gamma, \mathcal{S}_0 \rangle$. 
Let $\mathbf{w}^t$ be the robot state which consists of the robot's position $(p_x, p_y)$, velocity $(v_x, v_y)$, goal position $(g_x, g_y)$, and heading angle $\theta$. Let $\mathbf{h}_i^t$ be the current state of the $i$-th human at time $t$, which consists of the human's position and velocity $(p_x^i, p_y^i, v_x^i, v_y^i)$. Let $\mathbf{o}^t$ be the current observations of the static obstacles and walls. We define the state $s_t \in \mathcal{S}$ of the MDP as $s_t=[\mathbf{w}^t, \mathbf{o}^t, \mathbf{h}^t_1, ..., \mathbf{h}^t_n]$ if a total number of $n$ humans are observed at the timestep $t$, where $n$ may change within a range in different timesteps.


In each episode, the robot begins at an initial state $s_0\in \mathcal{S}_0$. According to its policy $\pi(a_t|s_t)$, the robot takes an action $a_t\in\mathcal{A}$ at each timestep $t$. In return, the robot receives a reward $r_t$ and transits to the next state $s_{t+1}$ according to an unknown state transition $\mathcal{P}(\cdot|s_t, a_t)$. 
Meanwhile, all humans also take actions according to their policies. 
The process continues until the robot reaches its goal, $t$ exceeds the maximum episode length $T$, or the robot collides with humans or obastacles. 

\subsection{Reward function}
\label{sec:reward}
The reward function awards the robot for reaching its goal and penalizes the robot for collisions with or getting too close to humans or obstacles. In addition, we add a potential-based reward shaping to guide the robot to approach the goal:
\begin{equation}
\label{eq:reward}
\begin{split}
\begin{gathered}
    r(s_t, a_t)  = 
        \begin{cases}
            -20, & \text{if } d_{min}^t < 0\\
            2(d_{min}^t - 0.25), & \text{if } 0<d_{min}^t<0.25\\
            20, & \text{if } d_{goal}^t \leq \rho_{robot}\\
            2(-d_{goal}^t+d_{goal}^{t-1}), & \text{otherwise}.
        \end{cases}
\end{gathered}
\end{split}
\end{equation}
where $d_{min}^t$ is the minimum distance between the robot and any human or obstacle at time $t$, and $d_{goal}^t$ is the $L2$ distance between the robot and its goal at time $t$. 
Intuitively, the robot gets a high reward when it approaches the goal while maintaining a safe distance from dynamic and static obstacles. 

\begin{figure}
\centering
\includegraphics[width=\columnwidth]{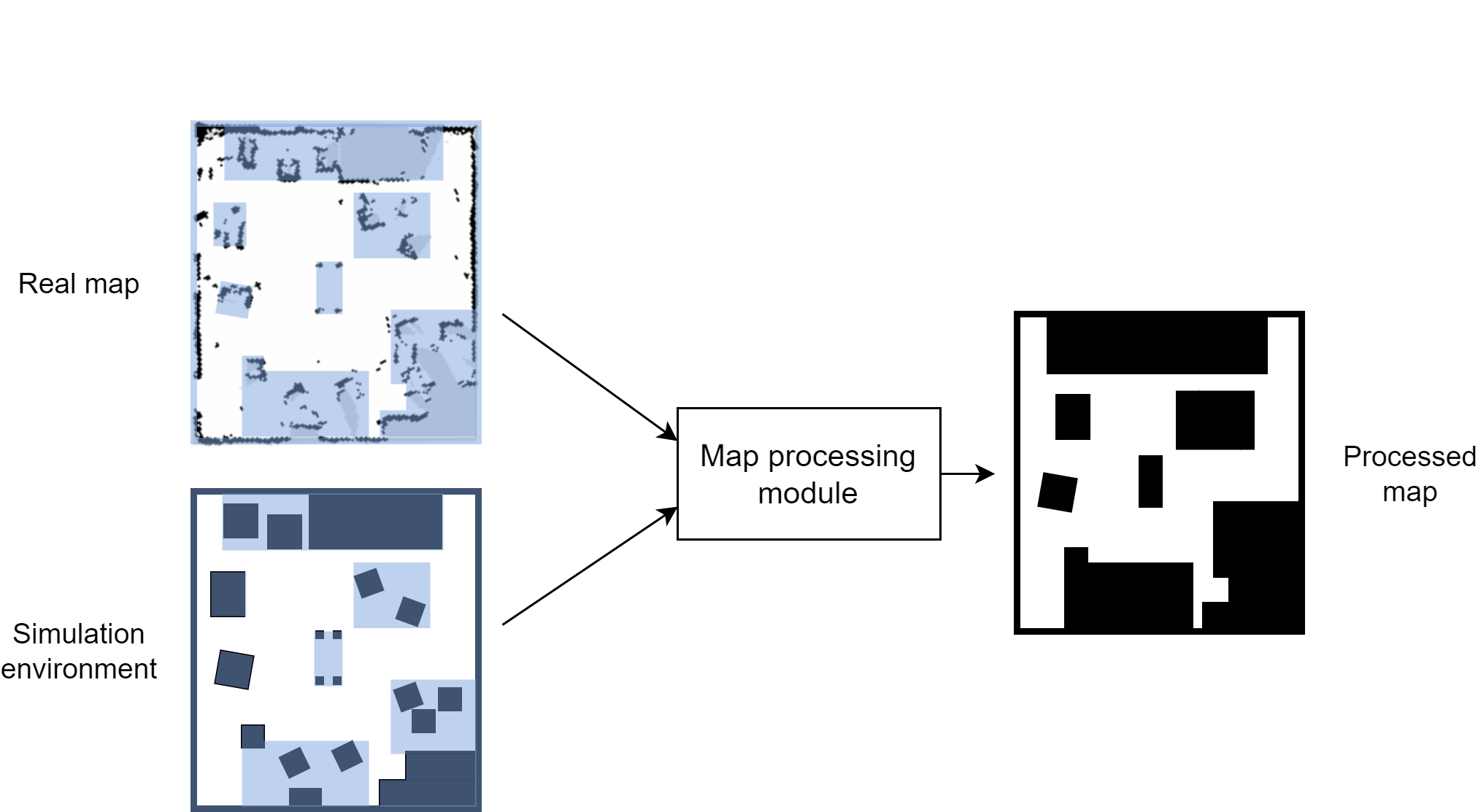} 
\caption{\textbf{Illustration of map processing.} Using off-the-shelf map processing techniques~\cite{liu2014extracting}, we can combine and smooth the edges of obstacles with irregular shapes. As a result, the processed map produces the obstacle point cloud representation, which introduces very small sim2real gaps. In the two raw maps on the left, we overlay the processed map on top of them for visualization purposes.}
\label{fig:map_process}
\vspace{-15pt}
\end{figure}
\section{Methodology}
\label{sec:method}
\begin{figure*}[ht]
\centering
\includegraphics[scale=0.5]{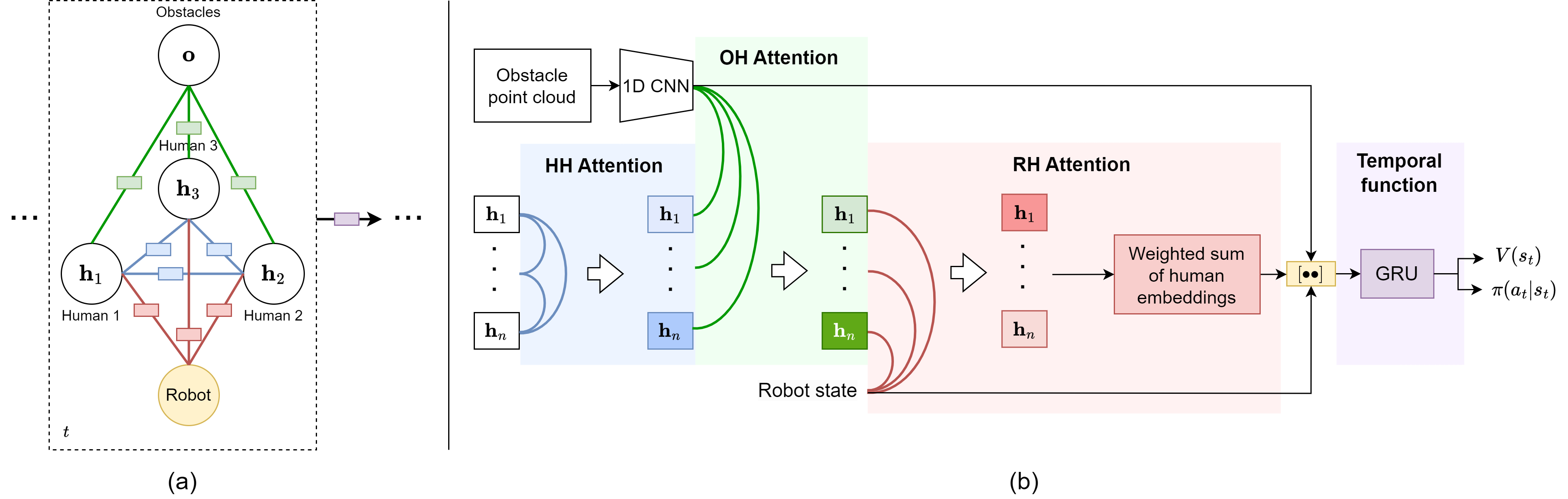}
\vspace{-5pt}
\caption{\textbf{The spatial-temporal interaction graph and the network architecture.} (a) Graph representation of crowd navigation. The robot node is in yellow, the $i$-th human node is $\mathrm{u}_i$, and the obstacle node is $o$. HH edges and HH functions are in blue, OH edges and OH functions are in green, and RH edges and RH functions are in red. The temporal function is in purple. (b) Our network. Three attention mechanisms are used to model the human-human, human-obstacle, and robot-human interactions. We use a GRU as the temporal function. }
\label{fig:st}
\vspace{-15pt}
\end{figure*}

\subsection{Scene representation}
High-dimensional raw sensor representation suffer from large sim2real gaps due to the presence of humans and complex obstacles and landscapes. 
Instead of investing in expensive high-fidelity simulators or laborious dataset collection, we use low-fidelity simulators. To circumvent sim2real gaps, our scene representation leverages processed information from perception, maps, and robot localization, which are relatively easier to obtain and robust to domain shifts. 

As shown in Fig.~\ref{fig:scene_rep}, at each timestep $t$,   we split a dynamic scene into a human representation denoted as $\mathbf{h}^t_1, ..., \mathbf{h}^t_n$, as well as an obstacle and constraint representation denoted as $\mathbf{o}^t$. 
In human representation, the position and velocity of each human is detected from off-the-shelf human detectors~\cite{Jia2020DRSPAAM,yolov3,deepsort}. By representing each human as a low-dimentional state vector, we abstract away detailed information such as gaits and appearance, which are difficult to model accurately~\cite{tsoi2022sean2,mavrogiannis2023core}. 
To obtain obstacle representation, as shown in Fig.~\ref{fig:map_process}, we first map the environment and process the map by combining close-by obstacles and approximating obstacle shapes as polygons. The map processing step is shown from Fig.~\ref{fig:scene_rep}(a) to Fig.~\ref{fig:scene_rep}(c). 
During navigation, assuming robot localization is available, we can compute a ``fake'' point cloud by performing a ray tracing algorithm centered at the robot location, as shown in Fig.~\ref{fig:scene_rep}(c). The ``fake'' point cloud contains approximate information on obstacles and constraints, which is sufficient for robot navigation. It is worth noting that compared to real point clouds from sensors, our obstacle representation is not affected by the presence of humans and is less sensitive to inaccuracies simulations of object appearance or shapes. 


\subsection{Structured interaction graph} 
Interactions among different entities contain essential information for multi-agent problems~\cite{vemula2018social,huang2019stgat, chen2019crowd, liu2020decentralized,chakraborty2023saber}.
We formulate constrained crowd navigation as a spatio-temporal (st) graph, which breaks the problem into smaller components in a structured fashion~\cite{jain2016structural}. 
In Fig.~\ref{fig:st}(a), the robot, all observed humans, and the observed static environment are nodes in the st-graph $\mathcal{G}_t$. At each timestep $t$, the edges that connect different nodes denote the spatial interactions among nodes. 
Different interactions have different effects on robot decision-making. 
Specifically, since we have control of the robot but not the humans, robot-human interactions have direct effects while human-human interactions have indirect effects on the robot actions. Since the agents are movable but the obstacles are static, interactions among agents are mutual while the influence of static obstacles on agents is one-way.
Thus, we categorize the spatial edges into three types: human-human (HH) edges (blue in Fig.~\ref{fig:st}), obstacle-human (OH) edges (green), and robot-human (RH) edges (red). 
The three types of edges allow us to factorize the spatial interactions into HH function, OH function, and RH function. Each function is a neural network that has learnable parameters.
Compared with the previous works that ignore some edges~\cite{chen2019crowd,liu2020decentralized,liu2023intention}, our method allows the robot to reason about all observed spatial interactions that exist in constrained crowded environments.


Since the movements of all agents cause the visibility of humans and obstacles to change dynamically, the set of nodes and edges and the parameters of the interaction functions may change correspondingly.
To this end, we integrate the temporal correlations of the graph $\mathcal{G}_t$ at different timesteps using another function denoted by the purple box in Fig.~\ref{fig:st}a. The temporal function connects the graphs at adjacent timesteps, which enables long-term decision-making of the robot.

The same type of edges share the same function parameters. This parameter sharing is important for the scalability of our st-graph because the number of parameters is kept constant with an increasing number of humans~\cite{jain2016structural}.

\subsection{Structured attention network}
In Fig.~\ref{fig:st}b, we derive our network architecture from the st-graph. We represent the HH, OH, and RH functions as feedforward networks with attention mechanisms, referred as HH attn, OH attn, and RH attn respectively. We represent the temporal function as a gated recurrent unit (GRU). We use $W$ and $f$ to denote trainable weights and fully connected layers.

\subsubsection{Attention mechanism}
The attention modules assign weights to all edges that connect to a node, allowing the node to attend to important edges or interactions. The 3 attention networks are similar to the scaled dot-product attention~\cite{vaswani2017attention}, which computes attention score using a query $Q$ and a key $K$, and applies the normalized score to a value $V$. 

\begin{equation} \label{eq:single_head_attn}
    \textrm{Attn}(Q, K, V) = \textrm{softmax}\left(\frac{QK^\top}{\sqrt{d}}\right)V
\end{equation}
where $d$ is the dimension of the queries and keys.

In HH attention, the current states of humans are concatenated and passed through linear layers to obtain $Q_{HH}^t ,K_{HH}^t, V_{HH}^t\in \mathbb{R}^{n \times d_{HH}}$, where $d_{HH}$ is the attention size for the HH attention.
\begin{equation}
    \begin{split}
        Q_{HH}^t = [\mathbf{u}^{t}_1, ...,\mathbf{u}^{t}_n]^\top W_{HH}^{Q} \\
        K_{HH}^t = [\mathbf{u}^{t}_1, ..., \mathbf{u}^{t}_n]^\top W_{HH}^{K} \\
        V_{HH}^t = [\mathbf{u}^{t}_1, ..., \mathbf{u}^{t}_n]^\top W_{HH}^{V}
    \end{split}
\end{equation}
We obtain the human embeddings $v_{HH}^t\in \mathbb{R}^{n \times d_{HH}}$ from Eq.~\ref{eq:single_head_attn}, and the number of attention heads is $8$.


In OH attention, the obstacle point cloud is fed into a 1D CNN, which outputs an obstacle embedding: $K_{OH}^t = f_{CNN}(\textbf{o}^t)$, where $K_{OH}^t\in \mathbb{R}^{1 \times d_{OH}}$.
$Q_{RH}^t, V_{RH}^t\in \mathbb{R}^{n \times d_{RH}}$ are linear embeddings of the weighted human features from HH attention $v_{HH}^t$. 
\begin{equation}
    \begin{split}
Q_{OH}^t = v_{HH}^t W_{OH}^{Q},\:
K_{OH}^t = v^t_O W_{OH}^{K},\:
V_{OH}^t = v_{HH}^t W_{OH}^{V}
 \end{split}
\end{equation}
We compute the attention score from $Q_{OH}^t$, $K_{OH}^t$, and $V_{OH}^t$ to obtain the twice weighted human features $v_{OH}^t\in \mathbb{R}^{1 \times d_{OH}}$ as in Eq.~\ref{eq:single_head_attn}. The number of attention heads is 1.

Similarly, in RH attention, we first embed the robot with a linear layer: $K_{RH}^t=f_{R}(\mathbf{w}^t)$, where $K_{RH}^t\in \mathbb{R}^{1 \times d_{RH}}$. 
$Q_{RH}^t, V_{RH}^t\in \mathbb{R}^{n \times d_{RH}}$ are linear embeddings of the weighted human features from OH attention $v_{OH}^t$. 
\begin{equation}
    \begin{split}
Q_{RH}^t = v_{OH}^t W_{RH}^{Q},\:
K_{RH}^t = \mathbf{w}^t W_{RH}^{K},\:
V_{RH}^t = v_{OH}^t W_{RH}^{V}
 \end{split}
\end{equation}
We compute the attention score from $Q_{RH}^t$, $K_{RH}^t$, and $V_{RH}^t$ to obtain the weighted human features for the third time $v_{RH}^t\in \mathbb{R}^{1 \times d_{RH}}$ as in Eq.~\ref{eq:single_head_attn}. The number of attention heads is 1.


In all three attention networks, we use binary masks that indicate the visibility of each human to prevent attention to invisible humans. The masks provide unbiased gradients to the networks, which stabilizes and accelerates the training.
\subsubsection{GRU}
We concatenate the robot embedding, the obstacle embedding, and the weighted human features and fed them into the GRU.
Finally, the hidden state of the GRU is input to a fully connected layer to obtain the value $V(s_t)$ and the policy $\pi(a_t|s_t)$.
We train the entire network with Proximal Policy Optimization (PPO)~\cite{schulman2017proximal}. 
\section{Simulation Experiments}
\label{sec:sim_exp}
\subsection{Simulation environment}
\begin{figure}[t]
\centering
\includegraphics[width=\columnwidth]{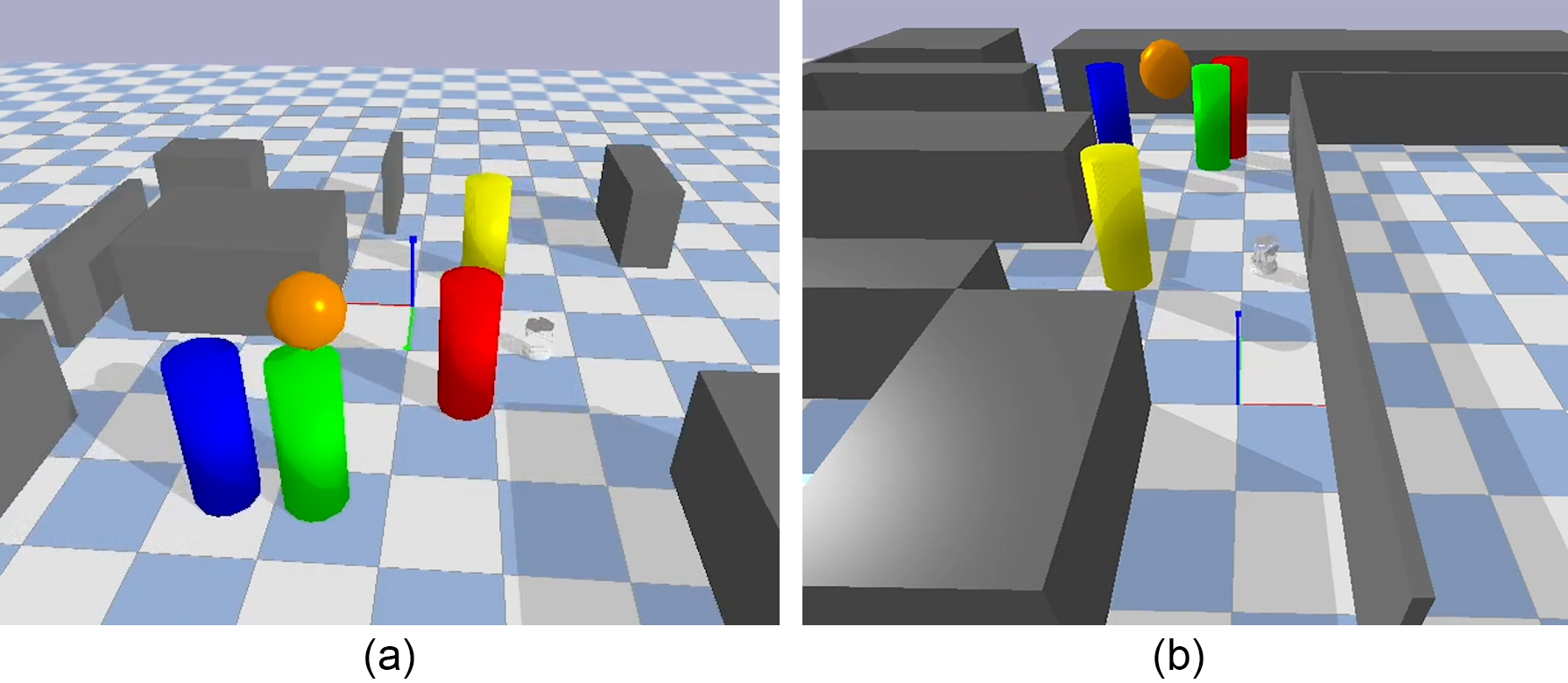}
\caption{\textbf{Two PyBullet simulation scenarios.} (a) Random environment with random obstacles and circle-crossing humans. (b) Sim2real environment with fixed obstacles and the random human flow is designed based on the layout.}
\label{fig:simulator}
\vspace{-15pt}
\end{figure}

Developed with PyBullet~\cite{coumans2019}, our simulator consists of two scenarios as shown in Fig.~\ref{fig:simulator}. 
We conduct simulation experiments in \textit{random environment} in Fig.~\ref{fig:simulator}(a). In each episode, obstacles are initialized with random shapes and random poses. The initial positions of the humans and the robot are also randomized. The human goals are set on the opposite side of their initial positions so that they cross each other in a circle. The number of humans varies from 2 to 4 and the number of obstacles varies from 7 to 9.

To simulate a continuous human flow, humans will move to new random goals immediately after they arrive at their goal positions.
All humans are controlled by ORCA~\cite{van2011reciprocal}.  80\% of humans do not react to the robot and 20\% of humans react to the robot. This mixed setting prevents our network from
learning an extremely aggressive policy in which the robot
forces all humans to yield while achieving a high reward, while maintaining a enough number of reactive humans to resemble the real crowd behaviors. 

We use unicycle kinematics for the robot.  
The action of the robot consists of desired translational and rotational accelerations $a_t = [a_{trans}, a_{rot}]$. The robot action space is discrete: the translational acceleration $a_{trans}\in\{-0.05\,m/s^2, 0\,m/s^2, 0.05\,m/s^2\}$ and the rotational acceleration $a_{rot}\in\{-0.1\,rad/s^2, 0\,rad/s^2, 0.1\,rad/s^2\}$. 
The translational velocity is clipped within $[0\, m/s, 0.5\, m/s]$ and rotational velocity is within $[-1\, rad/s, 1\,rad/s]$.
The robot motion is governed by the dynamics of TurtleBot 2i. 
We use holonomic kinematics for humans. 
The speed of humans is limited to 0.5 m/s to accommodate the speed of the robot. 

\subsection{Experiment setup}

\subsubsection{Baselines and ablation models}
To validate the effectiveness of the proposed scene representation, we compare our method with the following baselines:
\begin{itemize}
    \item Object-centric: The human representation is the same as our method, while obstacles are represented by the coordinates of their polygon vertices. The obstacle embedding network is a multi-layer perception instead of 1D CNN.
        \begin{table}[ht]
  \begin{center}
    \caption{Navigation results in random environment.}
    \label{tab:sim_results}
    \begin{tabular}{l c c c c} 
    \toprule
     \textbf{Method} & \textbf{Success}$\uparrow$ & \textbf{Collision}$\downarrow$ & \textbf{Timeout}$\downarrow$ & \textbf{Time}$\downarrow$\\
     
     \midrule
     
     Object-centric & 0.63 & 0.27 & 0.09 &12.02 \\
      
      Raw point cloud & 0.58 & 0.27 & 0.15 & 12.02 \\
      \midrule
      Ours, RH & 0.56 & 0.28 & 0.16 & 12.02\\
      
      Ours, RH+OH & 0.68 & 0.26 & 0.06 & \textbf{11.28}\\
      
      Ours, RH+HH+OH & \textbf{0.75} & \textbf{0.15} & \textbf{0.10} & 11.75 \\

      \bottomrule
    \end{tabular}
  \end{center}
  \vspace{-10pt}
\end{table}
    \item Raw point cloud: The humans and obstacles are represented as raw point clouds from a LiDAR. No human detection or fake point cloud is available. The point cloud is fed into a 1D CNN. 
    No attention network is used since all entities are mixed as a single point cloud. 
\end{itemize}

To validate the effectiveness of the structured graph network, we experiment with ablations of different attention models to justify the effect of 3 types of spatial interaction networks. Everything else except the presence of attention network(s) is kept the same as our method.
\begin{itemize}
    \item Ours, RH: The network has only RH attention and does not have HH or OH attention, similar to previous works such as \cite{chen2019crowd,leurent2019social,liu2020decentralized}.
    \item Ours, RH+OH: The network has only RH and OH attention and does not have HH attention.
    \item Ours, RH+HH+OH: The full version of our proposed network with all 3 attention modules.
\end{itemize}

\subsubsection{Training}
The policy is trained for $6\times 10^7$ steps in total. We run $16$ parallel environments to collect the robot's experiences. The learning rate is $8\times10^{-5}$ and decays linearly.
\subsubsection{Evaluation metrics}
We test all methods with $500$ random unseen test cases. Our metrics include success rate (Success), collision rate (Collision), timeout rate (Timeout), and average navigation time (NT) in seconds.

     
     
      
      


     
     
      
      


\subsection{Results}
In Table~\ref{tab:sim_results} and the \href{https://sites.google.com/view/constrained-crowdnav/home#h.2g2udt88bspg}{videos}, among the baselines, raw point cloud performs the worst because the network needs to extract both human and obstacle features from raw sensor inputs, which slows down the convergence. 
Object-centric achieves better results due to the presence of low-level state information but is outperformed by our method, because the vertex representation is sparse and not always useful for navigation. 
For example, among all vertices of a long and thin wall, the occluded vertices or the vertices outside of the robot field-of-view are fed into the network, yet none of them affects navigation. In addition, the vertex representation is sensitive to the permutation order of vertices, which poses extra challenges for learning~\cite{charles2017pointnet}. 
In contrast, our representation leverages low-level human states from detectors, which accelerates the convergence and thus achieves better performance with the same amount of training budget. For the obstacle representation, the ``fake'' point cloud is denser and more relevant to robot decision-making compared with vertex representation.

Among ablated models, we observe that if we remove HH attention, the success rate drops by 7\% because the interactions among humans are dense due to the presence of obstacles and environmental constraints. As a result, a human constantly changes its trajectories due to other humans. If we further remove OH attention, the success rate drops by another 12\% because the obstacles limit the traversable regions of agents, and thus human trajectories are also directly affected by obstacles. 
Thus, we conclude that reasoning about both human-human and human-robot interactions plays an important role in robot collision avoidance, in addition to robot-human interactions from previous works.

\section{Real-world Experiments}
\label{sec:real_exp}
We train our method in \textit{sim2real environment} in Fig.~\ref{fig:simulator}(b) and transfer the policy to a TurtleBot 2i in a real constrained indoor environment with pedestrians in a university building. We define 1 to 3 of human routes and robot routes based on the environment layout and randomly choose a route for each agent in each episode. In addition, 1 to 3 static humans are added. 
The poses of 11 obstacles are fixed. 

We use an Intel RealSense tracking camera T265 to obtain the pose of the robot. With an RPLIDAR-A3 laser scanner, we first remove non-human obstacles on a map, and then use a 2D LIDAR people detector~\cite{Jia2020DRSPAAM} to estimate the positions of humans.
From results (see \href{https://sites.google.com/view/constrained-crowdnav/home}{videos}), we observe that the robot can achieve goals without collisions no matter the pedestrians react to the robot or ignore the robot. However, the robot fails if the pedestrians intentionally block its path, since this kind of adversarial behavior is not simulated during training.
\begin{figure}[t]
\centering
\includegraphics[width=\columnwidth]{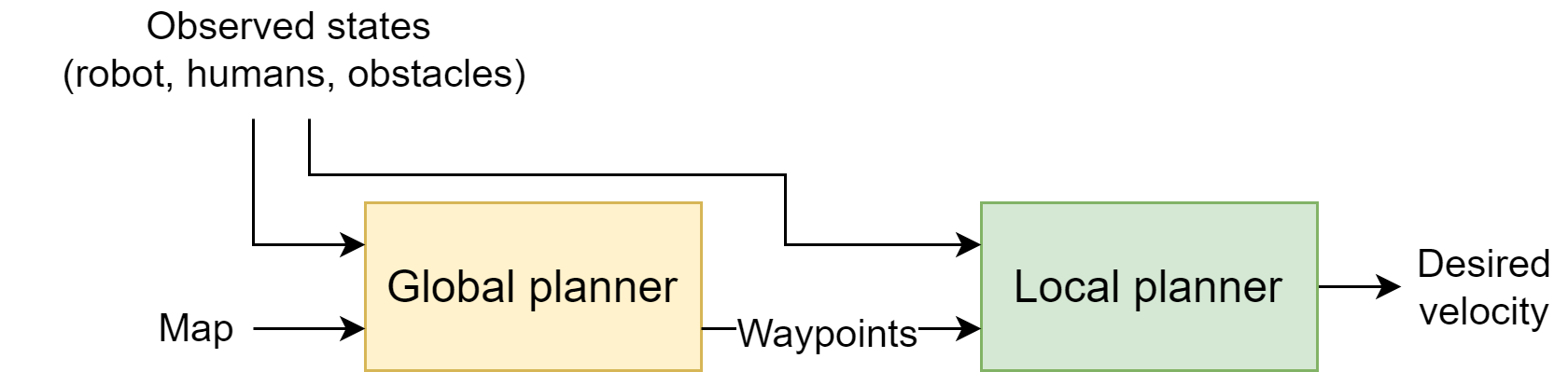} 
\caption{\textbf{A two-level hierarchical planner.} To enable long-horizon navigation, we can treat our method as a local planner and combine it with a global planner.}
\label{fig:hierarchical_planner}
\vspace{-15pt}
\end{figure}
\section{Conclusion}
In conclusion, to enable robots to navigate in constrained crowded environments, we propose a structured scene representation from preprocessed information. Then, we formulate the scenario as a st-graph, which leads to the derivation of a robot policy network that reasons about different interactions during navigation. We train the network with RL and demonstrate good results in both simulation and the real world.   
Our work encompasses the following limitations, which opens up opportunities for future work:
\begin{enumerate}
    \item The robot only achieves a 75\% success rate when it only navigates for 3 to 4 meters from start to goal. The task horizon and success rate are not enough for robot deployment in real applications such as last-mile delivery. To address this issue, as shown in Fig.~\ref{fig:hierarchical_planner}, we plan to adopt a hierarchical planner, which consists of a global planner that outputs waypoints, and a local planner that takes waypoints and performs low-level control. The current model can be used as the local planner. Possible options for the global planner include sample-based planners such as $A^*$ and RRT and learning-based policies. 
    \item As Sec.~\ref{sec:real_exp} discussed, the robot policy is overfitted to the simulated human behaviors. However, the simulated humans do not capture all the nuanced behavior patterns of real humans. A more realistic human motion model is necessary, which might need to be learned from real pedestrian data~\cite{chen2022combining}. In addition, adversarial RL training may also improve the robustness of the robot policy with respect to changes in human behaviors~\cite{pattanaik2017robust, pinto2017robust}.
\end{enumerate}



\newpage
\clearpage
\bibliographystyle{plainnat}
\bibliography{references}

\end{document}